\documentclass[pdflatex,sn-nature]{sn-jnl}

\usepackage{graphicx}%
\usepackage{subcaption}%
\usepackage{multirow}%
\usepackage{amsmath,amssymb,amsfonts}%
\usepackage{amsthm}%
\usepackage{mathrsfs}%
\usepackage[title]{appendix}%
\usepackage{xcolor}%
\usepackage{textcomp}%
\usepackage{manyfoot}%
\usepackage{booktabs}%
\usepackage{algorithm}%
\usepackage{algorithmicx}%
\usepackage{algpseudocode}%
\usepackage{listings}%
\usepackage{bm}
\usepackage{geometry}
\usepackage{adjustbox}
\usepackage{setspace}
\theoremstyle{thmstyleone}%
\usepackage[percent]{overpic}
\theoremstyle{thmstyletwo}%

\theoremstyle{thmstylethree}%

\begin{document}

\title[Article Title]{Higher-Order Geometric Updates for Levenberg--Marquardt Method via Riemann Normal Coordinates}

\author*[1,2]{\fnm{Jianing} \sur{Liu}}\email{liujn1218@mail.ustc.edu.cn}

\author*[2,3,4]{\fnm{Dong H.} \sur{Zhang}}\email{zhangdh@dicp.ac.cn}

\affil[1]{\orgdiv{State Key Laboratory of Chemical Reaction
Dynamics and Department of Chemical Physics}, \orgname{University of
Science and Technology of China}, \orgaddress{ \city{Hefei}, \postcode{230026}, \country{China}}}

\affil[2]{\orgdiv{State Key Laboratory of Chemical Reaction Dynamics}, \orgname{Dalian Institute of Chemical Physics, Chinese Academy of Sciences}, \orgaddress{\city{Dalian}, \postcode{116023}, \country{China}}}

\affil[3]{\orgname{University of Chinese Academy of Sciences}, \orgaddress{\city{Beijing}, \postcode{100049}, \country{China}}}

\affil[4]{\orgname{Hefei National Laboratory}, \orgaddress{ \city{Hefei}, \postcode{230088}, \country{China}}}

\abstract{Nonlinear least-squares objectives form the foundation of scientific machine-learning tasks, yet even curvature-aware optimizers remain geometrically inconsistent at finite step sizes: Levenberg–Marquardt (LM) derives its direction from local Riemannian metrics but realizes it as a straight parameter update. Here we introduce RNC-LM, which carries the LM direction along a locally constructed curved trajectory in Riemann normal coordinates. A recursive reformulation of the geodesic equation generates arbitrary finite-order corrections while reusing the same damped Gauss–Newton matrix factorization, and curve length is controlled separately from damping. On a reaction–diffusion physics-informed neural network benchmark, RNC-LM reduces relative $L^2$ errors below $8\times10^{-3}$, whereas L-BFGS, LM and LM with geodesic acceleration remain near one. On a large-scale machine-learning potential fitting task with 985,160 configurations, fourth-order RNC-LM reaches a fixed training-error target with a $34\times$ wall-clock speedup over LM. These results establish finite-step geometric realization as an distinct optimizer design principle.}
\maketitle

\section{Introduction}\label{sec1}

Models in modern machine learning and scientific computing are often highly nonlinear, with parameters serving primarily as coordinates for the underlying functions. Different parameterizations can represent the same model yet induce markedly different optimization trajectories. Optimizers should therefore respect the geometry of the underlying model rather than arbitrary parameter coordinates~\cite{bernstein2024arxiv, pmlr-v80-balles18a}. This principle motivates metric-aware optimization: natural-gradient methods derive their directions from the geometry of predictive distributions~\cite{Amari1998,JMLR:v21:17-678}, whereas Gauss--Newton and Levenberg--Marquardt (LM) methods use the local geometry induced by residual maps, with LM regularizing the Gauss--Newton system through damping~\cite{levenberg1944method, marquardt1963}.


However, a local metric defines only an infinitesimal descent direction $v$. The optimizer must still decide how this direction is realized as a finite parameter update, and a common choice is the coordinate straight-line update
\begin{equation}
    \theta_{p+1}=\theta_p+t v , 
\end{equation}
where $\theta_p$ denotes the parameter vector at iteration $p$ and $t>0$ is the step size. This update is a first order forward Euler discretization of the corresponding continuous gradient flow. However, it is not invariant under nonlinear reparameterization at finite step size: a straight line in one coordinate chart generally becomes curved in another. Thus, the resulting finite update remains parameterization-dependent, even when the tangent direction is geometrically well-defined.

This challenge has received long-standing attention within the nonlinear regression community. In the standard geometric picture of nonlinear least squares, the residual map $r(\theta)$ defines both the objective $C=\frac{1}{2}\|r(\theta)\|^2$ and a model manifold in data space, with the Jacobian $J$ inducing the pullback metric $J^\top J$~\cite{physrevlett.104.060201}. LM uses this metric to choose a descent direction, yet the finite update remains a straight step in the original coordinates. Bates and Watts formalized this mismatch by decomposing the nonlinearity into intrinsic curvature, which reflects the embedding of the model manifold, and parameter-effects curvature, which arises from the chosen parameterization~\cite{bates,bates1981parameter}. Kass later showed that zero parameter-effects curvature corresponds to a canonical parameterization, the Riemann normal coordinates, which can always be constructed locally at a regular point, even if a global zero-curvature parameterization does not exist in multiple dimensions~\cite{Robert_1984}. Transtrum and Sethna further showed that many complex nonlinear models are sloppy, with hyper-ribbon geometries whose Gauss--Newton or Fisher eigenvalues span many orders of magnitude. In such settings, parameter-effects curvature can dominate intrinsic curvature by orders of magnitude, making straight-line updates inefficient even when the underlying tangent direction is geometrically appropriate~\cite{physrevlett.104.060201,PhysRevE_83_036701}. Parameter-effects curvature is therefore not merely a concern of classical nonlinear regression; in modern scientific machine learning, the severe ill-conditioning of residual-based objectives such as PINN losses makes this curvature especially consequential at finite step size~\cite{NEURIPS2021_df438e52,pmlr-v235-rathore24a}.

This observation suggests that rather than redesigning the local quadratic subproblem, one can enhance LM by modifying how the tangent-space direction is realized as a finite update. In this spirit, Transtrum and Sethna introduced geodesic acceleration (LM-GA)~\cite{transtrum2012improvementslevenbergmarquardtalgorithmnonlinear}, which preserves the local LM direction while augmenting the coordinate straight-line update with a second-order correction. The correction is chosen to make the trajectory satisfy the geodesic equation at the expansion point, thereby reducing the leading-order effect of parameter-effects curvature. A related finite-step loss of reparameterization consistency has been studied for natural-gradient methods, where higher-order integrators better preserve the underlying invariant flow~\cite{pmlr-v80-song18a}. Brooks subsequently derived higher-order corrections from a path along which the residual vector decays linearly; this construction recovers geodesic acceleration at second order~\cite{brooks2024higherordercorrectionsoptimisersbased}. LM-GA and Brooks's residual-linear recursion nevertheless leave distinct geometric effects unresolved: the former corrects only the leading geodesic defect at the expansion point, whereas the latter does not explicitly track how the local tangent-space changes along the path, allowing lower-order residual components that are initially orthogonal to the tangent space to acquire new tangential projections along the trajectory.


Motivated by this observation, we introduce RNC-LM, a Levenberg--Marquardt method based on Riemann normal coordinates. At each iteration, RNC-LM takes the LM direction as the initial tangent vector and expresses the corresponding finite update in Riemann normal coordinates, where the local damped Gauss-Newton metric is approximately Euclidean and geodesics appear as straight lines. This yields a curved trajectory in the original parameter space that follows the local geometry to a prescribed order. A residual-space reformulation of the geodesic equation yields a recursion for the expansion coefficients. All coefficient systems use the same damped Gauss--Newton matrix as the LM step, so a single factorization can be reused, with each additional order requiring only a Taylor-mode directional-derivative right-hand side generated by automatic differentiation and one back-substitution. Because the linearized LM model and the finite-order curve have distinct ranges of validity, RNC-LM controls the distance travelled along the curve separately from LM damping: it first shortens the current trajectory when a full step is unreliable, and increases the damping and reconstructs the curve only if no acceptable point can be found.


We evaluate RNC-LM on classical nonlinear least-squares benchmarks and modern scientific machine-learning problems, including PINN failure modes and large-scale potential-energy-surface fitting. On a reaction--diffusion PINN benchmark, RNC-LM attains relative $L^2$ error below $8\times10^{-3}$, whereas L-BFGS, LM and LM-GA remain near unity; on a 985,160-configuration machine-learning potential fitting task, fourth-order RNC-LM reaches a fixed training-error target with a $34\times$ wall-clock time speedup. These results show that finite-step geometric realization can be as critical as the choice of tangent-space direction, establishing finite-step geometry as an optimizer design principle. They further suggest that RNC-LM provides a powerful and practical extension of existing LM methods.

\section{Results}\label{sec2}

\subsection{Riemann-normal-coordinate Levenberg--Marquardt updates}

We introduce RNC--LM to address the mismatch between the local LM direction and its coordinate-dependent finite-step realization. At the current iteration $\theta_p$, standard LM computes the direction $v$ from the damped Gauss--Newton system~\cite{levenberg1944method, marquardt1963}
\begin{equation}
Gv=-J^\top r,
\qquad
G=g+\lambda D,
\label{eq:sLM_eq}
\end{equation}
where $g=J^\top J$ is the pullback of the Euclidean residual-space metric, $\lambda\geq0$ is the damping parameter and $D$ is the Marquardt matrix. The matrix $g$ defines the target geometry, whereas $\lambda D$ regularizes the linear system. Standard LM takes a finite step along the straight-line $\theta_p + tv$ along the original parameter coordinates.

To reduce this coordinate dependence, RNC-LM works in Riemann normal coordinates centered at $\theta_p$, exploiting the connection between geodesic parameterizations and zero parameter-effects curvature~\cite{Robert_1984}.
In these coordinates, the geodesic initialized by $v$ is represented by the radial line $y(t)=tv$, where $t$ is the curve parameter~\cite{10.1063/1.1724316, hatzinikitas2000}. Expressed in the original parameters, the geodesic is approximated by the order-$K$ expansion
\begin{equation}
\theta_K(t)
=
\theta_p+
\sum_{q=1}^{K}\frac{t^q}{q!}c_q,
\qquad c_1=v.
\label{eq:rnc_curve}
\end{equation}

As shown in the Methods, the geodesic equation can be written in terms of residuals exactly as 
\begin{equation}
    J(\theta(t))^\top\ddot r(\theta(t))=0;
\end{equation}
 this condition removes the tangential component of the residual acceleration. Given $c_1,\ldots,c_{n-1}$, let $b_n$ denote the order-$t^{n-2}$ defect generated by the truncated curve. Canceling this defect leads to a unified linear equation for all the coefficients:
\begin{equation}
Gc_n=-b_n,
\qquad n=2,\ldots,K.
\label{eq:rnc_recursion}
\end{equation}
All orders therefore reuse one factorization of $G$, with each additional
order requiring only the construction of one Taylor-mode directional-derivative right-hand side followed by one additional solve.

In this construction, the order-$K$ curve
approximates the exact geodesic with error $O(t^{K+1})$ when $\lambda\to0$. Thus, $K=1$ recovers standard LM straight-line update, $K=2$ recovers geodesic acceleration, and $K>2$ successively corrects higher-order parameterization-induced tangential bending. The two parameters play distinct roles: $t$ controls the distance travelled along the constructed trajectory, while $K$ determines the truncation accuracy. Their complementary roles are examined in the benchmarks below.

\subsection{Classical nonlinear least-squares benchmarks}
\label{sec:classical_benchmarks}

\subsubsection{Curve order and curve length resolve distinct finite-step failures}

We used the generalized Rosenbrock problem introduced by Transtrum and Sethna~\cite{PhysRevE_83_036701} as a controlled ablation of the RNC order $K$ and the finite curve parameter $t$. The objective is
\begin{equation}
    C(\theta)
    =
    \frac{1}{2}
    \left[
        \theta_1^2
        +
        A^2
        \left(
            \theta_2-\frac{\theta_1^n}{n}
        \right)^2
    \right].
    \label{eq:generalized_rosenbrock}
\end{equation}
The parameter $A$ sets the transverse stiffness of the valley and largely determines the anisotropy and condition number of the Gauss--Newton metric, whereas $n$ determines the polynomial order of the valley centreline, $\theta_2=\theta_1^n/n$, and hence the order of nonlinear bending that a finite update must represent. We set $A=10^6$, for which condition number $\kappa(g)\approx 4\times10^{12}$ near the initial valley point. This matches the order of magnitude of the condition number observed in our neural-network experiments and provides a controlled stress test of optimization under
extreme anisotropy. Following Transtrum and Sethna, we initialized
each problem in the valley at $\theta_0=(1,1/n)$, and the convergence criterion was defined by $C(\theta)<10^{-4}$. Within RNC-LM, all settings used the same damping and acceptance rules; only $K$ and the maximum curve search depth were varied. Starting from $t=1$, trial values were selected by an interpolating line search with the safeguard $t_{j+1}\geq0.3t_j$. We allowed zero to three line-search reductions, giving minimum possible curve parameters $t_{\min}\in\{1,0.3,0.09,0.027\}$. If no acceptable point was found within this budget, $\lambda$ was increased and the RNC curve was recomputed.

\begin{figure}[htbp]
    \centering
    \includegraphics[width=0.7\textwidth]{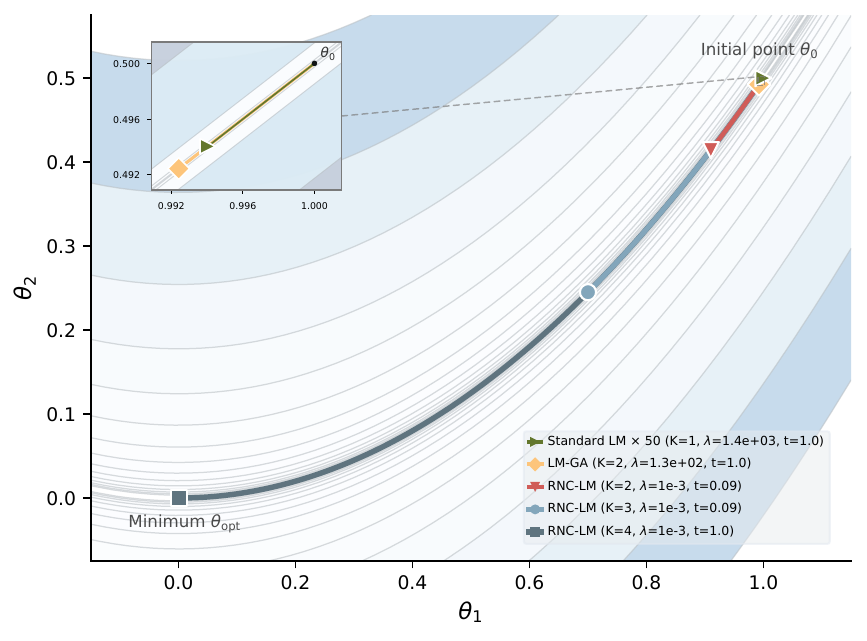}
    \caption{
First accepted trajectories from the common initial point on the generalized Rosenbrock objective. Contours show $\log_{10}{C(\theta)}$, and filled markers denote the accepted endpoints; labels report the corresponding $\lambda$ and $t$. The inset compares the much smaller LM and LM-GA updates near the initial point. For visibility, the LM
displacement vector in the inset is shown at \(50\) times its actual length; its direction is unchanged, and all other displacements are shown to scale.
}
    \label{fig:rosenbrock_trajectories}
\end{figure}

Figure~\ref{fig:rosenbrock_trajectories} shows the first accepted trajectories from the common initial point and provides two complementary comparisons.The RNC--LM curves with $K=2,3,4$ share the same damping parameter and initial tangent, so their differences reflect the effect of $K$ on the trajectory. LM-GA and second-order RNC--LM use the same second-order correction but respond differently to their finite-step control mechanism: LM-GA increases $\lambda$ and reconstructs a more strongly regularized curve, whereas RNC--LM retains the lower-damped curve and obtains acceptance by reducing $t$.

\begin{table}[htbp]
    \centering
    \caption{Iteration counts on two classical nonlinear least-squares benchmarks.}
    \label{tab:classical_benchmarks}

    \small
    \renewcommand{\arraystretch}{1.15}
    \setlength{\tabcolsep}{4.5pt}

    \begin{tabular*}{\linewidth}
        {@{\extracolsep{\fill}}lcccccccc@{}}
        \toprule
        \multicolumn{9}{c}{
            \textbf{(a) Generalized Rosenbrock, \(A=10^6\)}
        }\\
        \addlinespace[4pt]

        & \multicolumn{4}{c}{\(n=2\)}
        & \multicolumn{4}{c}{\(n=3\)} \\
        \cmidrule(lr){2-5}
        \cmidrule(lr){6-9}

        Method / \(t_{\min}\)
        & \(1.0\) & \(0.3\) & \(0.09\) & \(0.027\)
        & \(1.0\) & \(0.3\) & \(0.09\) & \(0.027\) \\
        \midrule

        LM
        & 8622 & --  & --  & --
        & 8069 & --  & --  & -- \\

        LM-GA
        & 152  & --  & --  & --
        & 183  & --  & --  & -- \\

        \addlinespace[2pt]

        RNC-LM (\(K=2\))
        & 152 & 191 & 187 & 11
        & 183 & 240 & 315 & 393 \\

        RNC-LM (\(K=3\))
        & 33 & 31 & 4 & 4
        & 33 & 41 & 7 & 7 \\

        RNC-LM (\(K=4\))
        & 2 & 2 & 2 & 2
        & 21 & 18 & 6 & 6 \\

        \bottomrule
    \end{tabular*}

    \vspace{10pt}

    \begin{tabular*}{\linewidth}
        {@{\extracolsep{\fill}}lccccccc@{}}
        \toprule
        \multicolumn{8}{c}{
            \textbf{(b) MGH10}
        }\\
        \addlinespace[4pt]

        & \multicolumn{1}{c}{Baseline}
        & \multicolumn{2}{c}{Second-order geodesic LM}
        & \multicolumn{4}{c}{RNC-LM with line search} \\
        \cmidrule(lr){2-2}
        \cmidrule(lr){3-4}
        \cmidrule(lr){5-8}

        Method
        & LM
        & Full step
        & Accel.\ ratio
        & 2nd
        & 3rd
        & 4th
        & 5th \\
        \midrule

        Iterations
        & 13576
        & 1735
        & 887
        & 286
        & 142
        & 76
        & 65 \\

        \bottomrule
    \end{tabular*}

    \vspace{6pt}

    \begin{minipage}{\linewidth}
        \footnotesize
        \raggedright
        Entries report outer iteration counts.
        \textbf{(a)} Under each \(n\), columns give the minimum curve parameter \(t_{\min}\) allowed in the RNC-LM curve search. LM and LM-GA do not use curve search and are reported only at $t_{\min}=1$.
        \textbf{(b)} ``Full step'' denotes the pure second-order geodesic correction without acceleration-ratio control or curve search. RNC-LM allows up to nine trial points per outer iteration.
    \end{minipage}
\end{table}
The first accepted trajectories provide a local illustration of the
distinct roles of \(K\) and \(t\). We quantified their consequences over
complete optimization runs by varying the expansion order and the available curve-search depth (Table~\ref{tab:classical_benchmarks}a). The results reveal two qualitatively different finite-step failures.

For the quadratic valley (\(n=2\)), RNC--LM ($K=2$) and LM-GA required 152
iterations with full steps, but only 11 iterations when the search could
reach \(t_{\min}=0.027\). The second-order trajectory was therefore
already geometrically useful, but the full displacement extended beyond
the range over which this local curve remained reliable. Reducing \(t\)
preserved the constructed curve while shortening the distance travelled
along it. At the same search depth, increasing the order further reduced
the iteration count to 4 for \(K=3\) and 2 for \(K=4\).

The cubic valley ($n=3$) showed the opposite pattern. Increasing the curve search budget did not rescue the second-order curve: the iteration count increased from 183 for full steps to 393 when the search could reach $t_\text{min}=0.027$. Although shortening $t$ can restore the local accuracy of a truncated curve, it cannot change the expansion order. Consequently, the optimizer is forced to take many short accepted steps. Increasing the expansion order instead produced a qualitative improvement. With sufficient line-search depth, the iteration count dropped to $7$ for $K=3$ and $6$ for $K=4$, indicating that the dominant benefit was obtained upon introducing the cubic term.

These results separate the roles of the two controls introduced by
RNC-LM. The curve parameter $t$ determines the distance travelled along
a constructed trajectory, whereas the order $K$ determines the
truncation accuracy of that trajectory. Backtracking is effective when
a useful curve is followed too far, while repeated acceptance of only
small values of $t$ indicates that a higher-order curve is required.

\subsubsection{Finite-step control and higher-order geometry on the MGH10 plateau}

We evaluated all 27 nonlinear-regression problems in the NIST StRD collection~\cite{nist_strd}, with the complete results reported in Supplementary Table~1. Here, we focus on MGH10, which is difficult for standard LM and was previously identified as a problem for which acceleration-ratio control can be particularly effective. Its model is
\begin{equation}
    f(x;\theta)
    =
    \theta_1
    \exp\left(
        \frac{\theta_2}{x+\theta_3}
    \right).
    \label{eq:mgh10_model}
\end{equation}
When $|\theta_3|\gg |x|$, the response is governed by the effective
parameter combinations
$\phi_0=\theta_1\exp(\theta_2/\theta_3)$ and
$\phi_k=(-1)^k\theta_2/\theta_3^{k+1}$ for $k\geq1$.
The problem exhibits pronounced parameter-effects curvature due to the strongly separated scales that make several original parameter directions nearly indistinguishable to first order, while producing non-negligible higher-order nonlinear residual changes. This produces a nearly rank-deficient plateau on which the finite-step realization of the LM direction becomes critical.

Starting from the first of NIST's two standard starting points, standard LM required 13,576 outer iterations. Retaining identical trust-region and damping controls, LM-GA reduced this count to 1,735. By disabling the acceleration safeguard and curve search, this comparison directly isolates the intrinsic contribution of the second-order trajectory.

We next compared two additional finite-step control mechanisms: the acceleration-ratio safeguard in LM-GA and the interpolating curve search in RNC-LM ($K=2$). LM-GA responds to an excessive acceleration ratio by increasing the damping and reconstructing the local update, whereas RNC-LM first retains the current second-order curve and searches for an acceptable distance $t$ along it.

The results are reported in Table~\ref{tab:classical_benchmarks}b. On MGH10, the acceleration-ratio safeguard reduced the iteration count to 887 with the acceleration-ratio threshold $\alpha=0.75$, confirming its effectiveness on this problem. Tightening the threshold to $\alpha=0.1$ resulted in 881 iterations,
yielding essentially no further improvement. The performance of LM-GA had therefore largely saturated with respect to $\alpha$, and its
remaining difference from RNC-LM cannot be attributed simply to an insufficiently stringent acceleration threshold.

By contrast, adaptive curve-length control reduced the iteration count to 286 for second-order RNC-LM. Thus, although both mechanisms improved over the LM-GA without additional control mechanisms, direct adaptation of the curve parameter was markedly more effective on the MGH10 plateau than acceleration-ratio control.

Increasing the RNC order produced a further monotonic improvement, reducing the iteration count to 142 for $K=3$, 76 for $K=4$ and 65 for $K=5$. This trend is consistent with the scale-separated effective parameters above, whose finite-step variation is represented progressively more accurately as the expansion order increases. The results therefore identify two distinct sources of improvement:
adaptive curve-length control is already beneficial at second order, whereas higher orders provide an additional gain by improving the geometric representation of the trajectory itself.


\subsection{Physics-informed neural networks}

We next evaluate RNC-LM on a modern scientific machine learning problem. We adopt physics-informed neural networks (PINNs) as the benchmark because their training objectives are naturally formulated as nonlinear least-squares problems. A typical PINN loss function is composed of PDE, boundary condition, and initial condition residuals:
\begin{equation}
    \mathcal{L}_{\mathrm{PINN}}(\theta)
    =
    w_{\mathrm{PDE}} \frac{\|r_{\mathrm{PDE}}(\theta)\|_2^2}{N_{\mathrm{PDE}}}
    +
    w_{\mathrm{IC}} \frac{\|r_{\mathrm{IC}}(\theta)\|_2^2}{N_{\mathrm{IC}}}
    +
    w_{\mathrm{BC}} \frac{\|r_{\mathrm{BC}}(\theta)\|_2^2}{N_{\mathrm{BC}}},
    \label{eq:pinn_objective}
\end{equation}
where for each component $q \in \{\mathrm{PDE}, \mathrm{IC}, \mathrm{BC}\}$, 
$r_q$ represents the residual vector evaluated at $N_q$ collocation points, 
and $w_q$ is its associated loss weight.

The resulting optimization problem is known to be challenging. The PDE residual 
$r_{\mathrm{PDE}}$ contains differential operators applied to the network output, 
and the different residual components can be severely imbalanced. Previous studies have shown that PINN failures are often caused not by limited network expressivity, but by ill-conditioning, gradient pathologies, and difficult loss landscapes induced by the PINN formulation itself~\cite{NEURIPS2021_df438e52, wang2022110768, pmlr-v235-rathore24a}. These difficulties have motivated the development and adoption of curvature-aware optimization methods for PINNs, including natural-gradient, LM and Newton--CG variants.

We consider the one-dimensional reaction--diffusion failure-mode benchmark~\cite{NEURIPS2021_df438e52}, given by
\begin{equation}
    \frac{\partial u }{\partial t} - \nu \frac{\partial^2 u}{\partial x^2} - \rho u (1-u) = 0, \qquad x\in [0,2\pi), \quad t \in (0,1],
\end{equation}
here, $\rho$ is the reaction coefficient and $\nu>0$ is the diffusion coefficient. We impose the periodic boundary conditions and Gaussian initial profile used in the original benchmark.

Direct PINN training over the full space--time domain for this equation can fail to recover the physical solution, even though the network architecture has sufficient expressive capacity. This failure is associated with the difficult optimization problem induced by imposing the governing equation as a soft residual constraint. Previous work has shown this failure can be reduced through curriculum regularization or sequence-to-sequence time marching. These and other related strategies, however, modify the training procedure~\cite{NEURIPS2021_df438e52,pmlr-v202-daw23a, fang2024ensemble}. Subsequent optimizer-focused studies have demonstrated substantial improvements on convection and other challenging PINN benchmarks~\cite{pmlr-v235-rathore24a,neurips2024_49f07f60}. However, to the best of our knowledge, this reaction--diffusion failure mode has not previously been overcome solely through optimizer design under a fixed single-network, full-domain PINN setup. 

With the reaction coefficient fixed at $\rho=5$, we evaluate $\nu=5$ and $6$, the settings from the original benchmark for which direct full-domain PINN training exhibits pronounced difficulty, and include $\nu=10$ and $20$ as stress tests beyond the original benchmark range. To assess robustness in the representative failure case $\nu=5$, we additionally conduct two paired studies, each comprising ten independent runs: one varies the network initialization while holding the collocation set fixed, and the other varies the collocation set while holding the network initialization fixed. All four optimizers--L-BFGS, standard LM, LM-GA and RNC-LM--use the same network architecture and residual formulation; within each paired run, they also share identical initial parameters and collocation points. All RNC-LM experiments in this section use $K=3$, and allow up to three curve searches per outer iteration. Full experimental details are provided in the Supplementary Information (SI).

Table~\ref{tab:pinn_reaction_diffusion} shows marked differences in performance among the optimizers on this reaction--diffusion problem. L-BFGS, standard LM
and LM-GA substantially reduce the training loss but fail to recover the physical solution at $\nu=5$ and $6$, retaining relative $L^2$ errors close to unity. RNC-LM, by contrast, reaches relative $L^2$ errors of order $10^{-3}$. This separation persists when $\nu$ is increased to $10$ and $20$ as more challenging stress tests: RNC-LM maintains relative $L^2$ errors of order $10^{-3}$, whereas all competing optimizers remain at large solution errors.
\begin{figure}[htbp]
    \centering

    \begin{subfigure}[t]{0.48\linewidth}
        \centering
        \begin{overpic}[width=\linewidth]
            {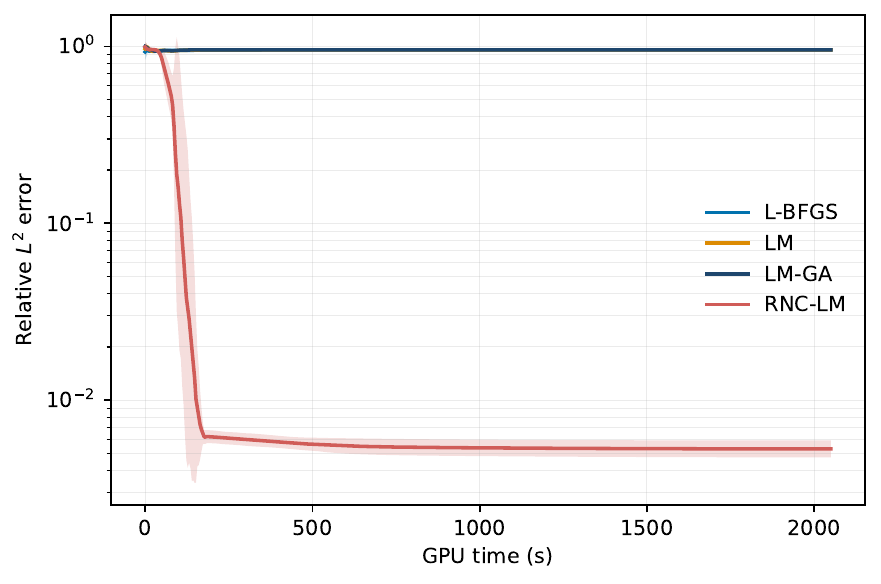}
            \put(5,67){(a)}
        \end{overpic}
        \phantomsubcaption
        \label{fig:pinn_rel_l2}
    \end{subfigure}
    \hfill
    \begin{subfigure}[t]{0.48\linewidth}
        \centering
        \begin{overpic}[width=\linewidth]
            {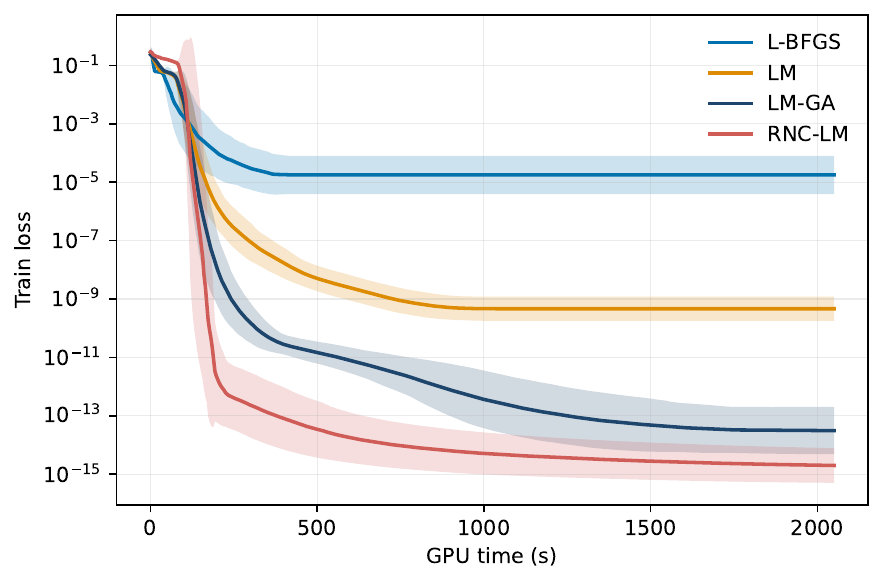}
            \put(5,67){(b)}
        \end{overpic}
        \phantomsubcaption
        \label{fig:pinn_train_loss}
    \end{subfigure}

    \medskip

    \begin{subfigure}[t]{0.48\linewidth}
        \centering
        \begin{overpic}[width=\linewidth]
            {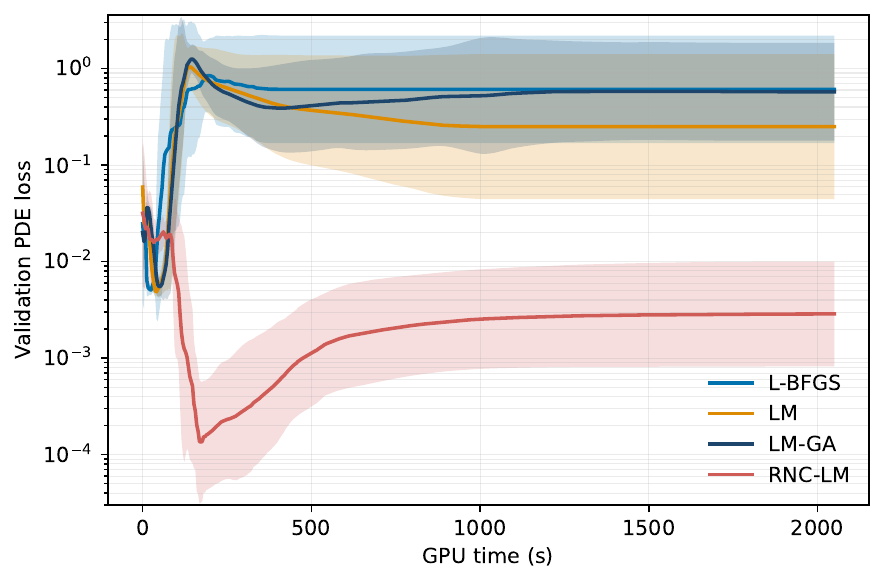}
            \put(5,67){(c)}
        \end{overpic}
        \phantomsubcaption
        \label{fig:pinn_validation_pde}
    \end{subfigure}
    \hfill
    \begin{subfigure}[t]{0.48\linewidth}
        \centering
        \begin{overpic}[width=\linewidth]
            {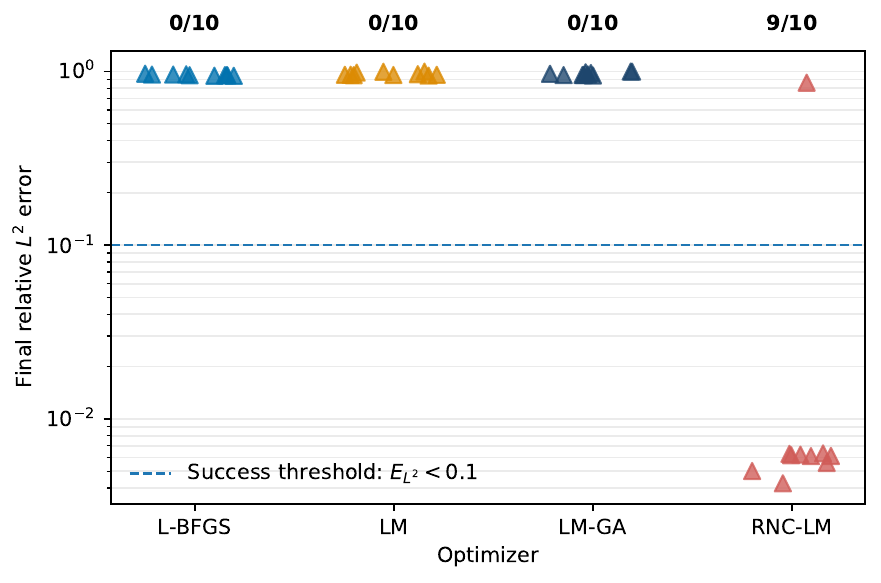}
            \put(5,67){(d)}
        \end{overpic}
        \phantomsubcaption
        \label{fig:pinn_collocation_robustness}
    \end{subfigure}

    \caption{\textbf{Robustness and residual generalization on the reaction--diffusion PINN benchmark at \(\rho=5\) and \(\nu=5\).} \textbf{a}, Mean relative \(L^2\) error as a function of wall-clock time, with shading indicating one standard deviation across ten independent network initializations. \textbf{b}, Corresponding PINN training loss. \textbf{c}, Validation PDE loss.
    \textbf{d}, Final relative \(L^2\) errors across ten independently sampled collocation sets with a fixed network initialization. Each point represents one paired run. The dashed line marks the success criterion \(<0.1\), and the numbers above each optimizer report the number of successful runs out of ten. Within each paired run, all optimizers use identical initial parameters and collocation points.}
    \label{fig:PINN_loss}
\end{figure}

We next examine the $\nu=5$ case in greater detail. First, we fix the collocation points and repeat over 10 independent runs. All four optimizers reduce the training objective, although to significantly different levels (Fig.~\ref{fig:PINN_loss}b). However, this reduction in training loss does not translate into physical accuracy. The validation PDE losses of L-BFGS, standard LM and LM-GA remain high after a brief initial drop, whereas RNC-LM reduces the validation loss by several 
additional orders of magnitude (Fig.~\ref{fig:PINN_loss}c). Correspondingly,
the competing optimizers remain at relative $L^2$ errors close to 1.0,
while RNC-LM consistently enters the low-error regime across the ten
initializations (Fig.~\ref{fig:PINN_loss}a).

Next, we test the sensitivity to the choice of training points by fixing 
the network initialization and randomly sampling the collocation set ten times. 
RNC-LM recovers the physical solution in $9/10$ runs
(Fig.~\ref{fig:PINN_loss}d). In the remaining run, it fails to enter the
accurate-solution regime, with a final relative $L^2$ error of $0.36$.
By contrast, L-BFGS, standard LM and LM-GA achieve success rates of
\(0/10\). RNC-LM therefore 
significantly increases the success rate of direct, full-domain PINN training 
under both random initializations and training point samples.

\begin{table}[htbp]
\centering
\caption{
Reaction--diffusion PINN results. For each value of \(\nu\), we report the final training loss and the relative \(L^2\) error with respect to the reference solution. RNC-LM achieves both the lowest residual loss and the lowest physical error across all tested regimes.
}
\label{tab:pinn_reaction_diffusion}
\renewcommand{\arraystretch}{1.2}
\setlength{\tabcolsep}{5pt}
\begin{tabular}{llcccc}
\toprule
\toprule
 & & L-BFGS & LM & LM-GA & RNC-LM \\
\midrule
\multirow{2}{*}{$\nu = 5$}
& Train Loss
& $2.27 \times 10^{-6}$
& $2.93 \times 10^{-10}$
& $9.84 \times 10^{-15}$
& \textbf{$4.43 \times 10^{-16}$} \\
& Relative \(L^2\)
& $9.57 \times 10^{-1}$
& $9.59 \times 10^{-1}$
& $9.57 \times 10^{-1}$
& \textbf{$6.56 \times 10^{-3}$} \\
\midrule
\multirow{2}{*}{$\nu = 6$}
& Train Loss
& $8.94 \times 10^{-6}$
& $3.22 \times 10^{-10}$
& $1.92 \times 10^{-13}$
& \textbf{$7.69 \times 10^{-16}$} \\
& Relative \(L^2\)
& $9.87 \times 10^{-1}$
& $9.82 \times 10^{-1}$
& $9.80 \times 10^{-1}$
& \textbf{$4.57 \times 10^{-3}$} \\
\midrule
\multirow{2}{*}{$\nu = 10$}
& Train Loss
& $1.67 \times 10^{-4}$
& $2.45 \times 10^{-10}$
& $2.92 \times 10^{-14}$
& \textbf{$1.31 \times 10^{-15}$} \\
& Relative \(L^2\)
& $9.98 \times 10^{-1}$
& $9.98 \times 10^{-1}$
& $9.97 \times 10^{-1}$
& \textbf{$7.87 \times 10^{-3}$} \\
\midrule
\multirow{2}{*}{$\nu = 20$}
& Train Loss
& $1.47 \times 10^{-4}$
& $8.95 \times 10^{-11}$
& $9.56 \times 10^{-15}$
& \textbf{$2.98 \times 10^{-16}$} \\
& Relative \(L^2\)
& $9.98 \times 10^{-1}$
& $9.98 \times 10^{-1}$
& $9.97 \times 10^{-1}$
& \textbf{$5.87 \times 10^{-3}$} \\
\bottomrule
\bottomrule
\end{tabular}
\end{table}

\begin{table*}[!t]
\centering
\caption{\textbf{Comparison of optimizer performance on three PINN benchmark equations.}
The benchmarks comprise the one-dimensional convection equation $u_t+\beta u_x=0$ with $\beta=40$, the reaction equation $u_t-\rho u(1-u)=0$ with $\rho=5$, and the wave equation $u_{tt}-4u_{xx}=0$ with initial-condition frequency parameter $\beta=5$; complete initial and boundary conditions are given in the SI. Entries report the final training loss and relative $L^2$ error (L2RE). Published values for Adam+L-BFGS and Adam+L-BFGS+NNCG are reproduced from Ref.~\cite{pmlr-v235-rathore24a}. Values for LM, LM-GA and RNC-LM are medians over ten independent runs with different random seeds; the
corresponding means and standard deviations are reported in
Supplementary Table~2. Boldface indicates the lowest reported value in
each column.}
\label{tab:optimizer_comparison}

\small
\setlength{\tabcolsep}{5pt}
\renewcommand{\arraystretch}{1.08}

\begin{adjustbox}{max width=\linewidth,center}

\begin{tabular}{@{}lcccccc@{}}
\toprule
\multirow{2}{*}{Optimizer}
& \multicolumn{2}{c}{Convection}
& \multicolumn{2}{c}{Reaction}
& \multicolumn{2}{c}{Wave} \\
\cmidrule(lr){2-3}
\cmidrule(lr){4-5}
\cmidrule(lr){6-7}
& Loss & L2RE
& Loss & L2RE
& Loss & L2RE \\
\midrule
Adam+L-BFGS
& \(5.95\times10^{-6}\) & \(4.19\times10^{-3}\)
& \(5.26\times10^{-6}\) & \(1.92\times10^{-2}\)
& \(1.12\times10^{-3}\) & \(5.52\times10^{-2}\) \\

Adam+L-BFGS+NNCG
& \(3.63\times10^{-7}\) & \(1.94\times10^{-3}\)
& \(2.89\times10^{-7}\) & \(9.92\times10^{-3}\)
& \(6.13\times10^{-5}\) & \(1.27\times10^{-2}\) \\

LM
& \(1.81\times10^{-8}\) & \(1.22\times10^{-4}\)
& \(2.66\times10^{-11}\) & \(2.00\times10^{-4}\)
& \(4.72\times10^{-6}\) & \(7.61\times10^{-3}\) \\

LM-GA
& \(1.71\times10^{-10}\) & \(1.23\times10^{-4}\)
& \(2.27\times10^{-16}\) & \(1.97\times10^{-7}\)
& \(2.01\times10^{-11}\) & \(3.27\times10^{-6}\) \\

RNC-LM
& \(\bm{2.64\times10^{-16}}\)
& \(\bm{1.71\times10^{-7}}\)
& \(\bm{1.32\times10^{-16}}\)
& \(\bm{1.73\times10^{-7}}\)
& \(\bm{5.47\times10^{-13}}\)
& \(\bm{1.51\times10^{-6}}\) \\
\bottomrule
\end{tabular}
\end{adjustbox}
\end{table*}

Having established that RNC-LM can overcome a known PINN optimization failure mode, we further compare it with representative PINN training strategies on convection, reaction, and wave equations, including the conventional Adam+L-BFGS pipeline, the recent NysNewton-CG (NNCG) method, standard LM, and LM-GA. As shown in Table~\ref{tab:optimizer_comparison}, RNC-LM achieves the lowest training loss and relative $L^2$ error on all three benchmarks. The improvement over LM-GA is most pronounced on the convection benchmark, another widely studied PINN failure mode~\cite{NEURIPS2021_df438e52}, while RNC-LM also provides further accuracy gains on the reaction and wave equations.


Taken together, the reaction--diffusion experiments support the view that this failure mode is primarily optimization-driven rather than a consequence of insufficient network expressivity. Although the competing methods can substantially reduce the loss at the training collocation points, this reduction does not necessarily translate into an accurate physical solution, whereas RNC-LM consistently enters the low-error regime without modifying the network architecture or training setup. The additional convection, reaction, and wave benchmarks further show that this advantage is not confined to a single pathological problem. More broadly, these findings suggest that the finite-step geometric update can be critical for traversing difficult PINN loss landscapes.


\subsection{Machine-Learning Potential-Energy-Surface Fitting}

Finally, we test whether the higher-order RNC construction remains computationally beneficial on a larger scientific regression problem. We consider high-accuracy Born--Oppenheimer potential-energy-surface fitting. This problem involves a large residual vector and an explicitly constructed Gauss-Newton matrix that nearly approaches the memory limit of a single A100 GPU. 

We use a fundamental-invariant neural-network (FI-NN) potential for three-body H$_2$O cluster, following the framework in Refs.~\cite{d5cs00104h, 10.1093/nsr/nwad321}. The dataset contains 985,160 configurations represented by 1000 FI descriptors. The neural network is a fully connected MLP with architecture $\{1000,55,55,55,1\}.$ We compare standard LM with RNC-LM ($K=3$) and ($K=4$). All methods use the same initialization and data split. RNC-LM allows up to three line-search trials per outer iteration.

\begin{figure}[htbp]
    \centering

    \begin{subfigure}{0.48\textwidth}
        \centering
        \includegraphics[width=\textwidth]{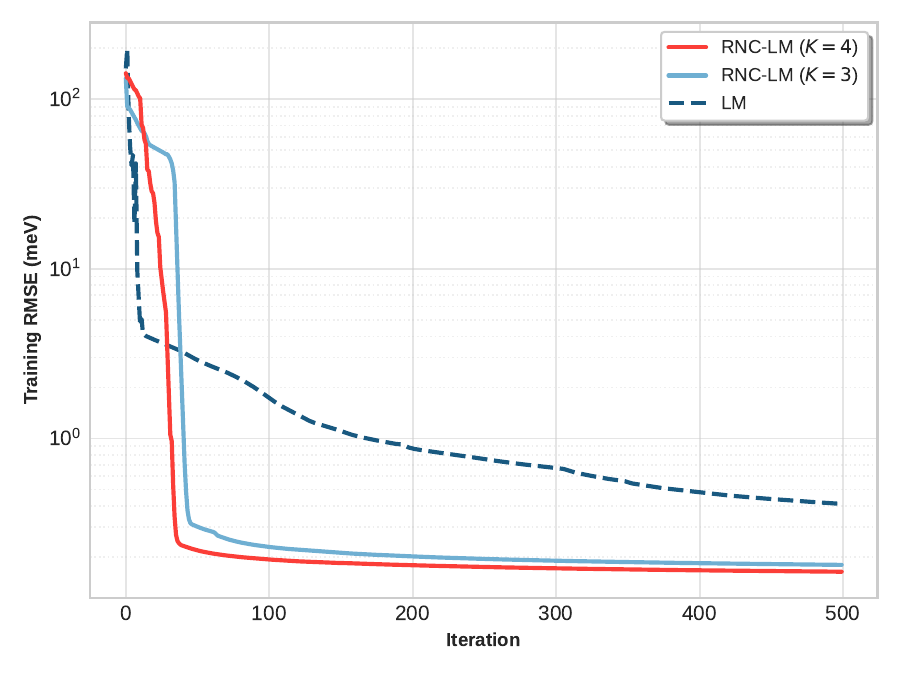}
        \caption{Training-set loss curve}
        \label{fig:train_loss}
    \end{subfigure}
    \hfill
    \begin{subfigure}{0.48\textwidth}
        \centering
        \includegraphics[width=\textwidth]{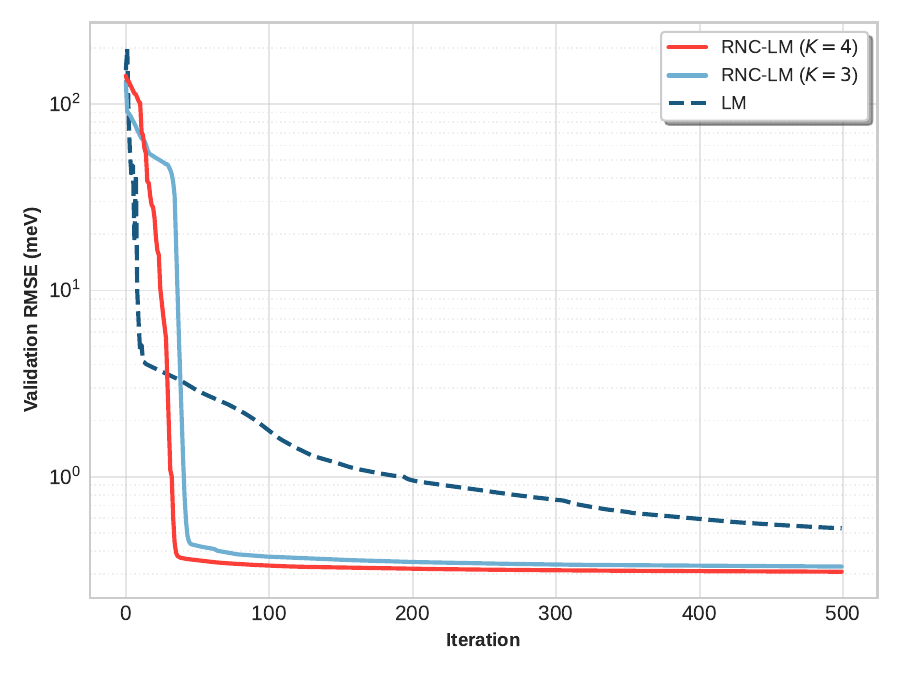}
        \caption{Validation-set loss curve}
        \label{fig:test_loss}
    \end{subfigure}

    \caption{
    Loss-curve comparison on the three-body \(\mathrm{H}_2\mathrm{O}\) potential-energy-surface fitting task.
    RNC-LM accelerates convergence on both the training and validation sets.
    }
    \label{fig:loss_comparison}
\end{figure}

Figure~\ref{fig:loss_comparison} compares the training- and validation-set RMSE trajectories. Both RNC-LM implementations reduce the RMSE in substantially fewer outer iterations than standard LM, with the $K=4$ implementation converging faster than the $K=3$ counterpart. Since the higher-order corrections also require additional directional derivatives and linear solves, we further compare the wall-clock time required to reach a common training-RMSE target of $0.186$ meV. Standard LM needs the entire 5,000-iteration budget and $387.28$ hours to reach this target, whereas RNC-LM ($K=3$) reaches the same target in 385 iterations and $30.24$ hours, corresponding to a $12.8\times$ speedup, and RNC-LM ($K=4$) requires only 144 iterations and $11.39$ hours, yielding a $34.0\times$ speedup. 

These results show that the benefit of RNC-LM is not limited to low-dimensional benchmarks. Even in a large-scale scientific machine-learning  regression task, higher-order RNC updates significantly reduce both the number of outer iterations and the total wall-clock time. This suggests that the proposed geometric correction can lower the computational cost of high-accuracy neural-network potential-energy-surface fitting and may help extend such potential energy surface (PES) models to larger molecular systems.

\section{Discussion}
Our results identify finite-step realization as a distinct design axis in curvature-aware optimization, complementary to the choice of local metric. LM-type methods use local geometric information to determine an infinitesimal search direction, but typically realize that direction through an additive, straight-line step in parameter coordinates. RNC-LM separates these two choices: it preserves the local LM subproblem while replacing the additive update with a finite-order geometric trajectory based on Riemann normal coordinates. The improvements observed across nonlinear-regression benchmarks, PINN failure modes, and large-scale potential fitting tasks show that, within the settings studied here, finite-step realization can substantially affect optimization even when the local direction is held fixed.

The present implementation explicitly constructs and factorizes the metric, limiting its direct applicability to settings in which these operations are tractable. However, the RNC recursion only requires repeated solutions of linear systems involving the same metric. This structure suggests two routes to scalable implementations: structured approximations of the metric, such as K-FAC~\cite{pmlr-v37-martens15, luk2018arxiv} and matrix-free Krylov-subspace methods, including preconditioned conjugate gradients and subspace recycling techniques~\cite{doi:10.1137/040607277, doi:10.1137/21m1466244}. Future work should characterize the trade-offs among truncation order, metric-approximation error, linear-solve accuracy, and computational cost.

The perspective developed here extends beyond LM and nonlinear least squares. Natural-gradient and stochastic-reconfiguration methods also determine descent directions from local metrics—the Fisher information and Fubini–Study metrics—but usually apply those directions through coordinate straight-line updates. The distinction between choosing a local direction and constructing its finite-step trajectory is therefore relevant on these manifolds as well, particularly when the metric varies appreciably along the step. Extending the RNC construction to these settings is a promising direction for future work. The practical implications are especially relevant to scientific machine learning, where PINNs, neural potentials, and other parameter-rich models can present severely ill-conditioned optimization problems. Our results suggest that finite-step geometric corrections offer a route to more reliable training without changing the model architecture or training pipelines. Further evaluation across algorithms and scales will clarify whether finite-step geometry can serve as a general principle of optimizer design.
\section{Methods}\label{sec11}
\subsection{Derivation of the geodesic equation in residual embedding} 

Let $r:\Theta\subset\mathbb R^N\rightarrow\mathbb R^M$ be a sufficiently smooth residual map, with full rank Jacobian $J_{m\mu}=\partial_\mu r_m$. The Euclidean metric in residual space induces the Gauss--Newton pullback metric
\begin{equation}
    g_{\mu\nu}(\theta)=J_{m\mu}(\theta)J_{m\nu}(\theta)
\end{equation}
on parameter space. Greek indices label parameter components, Latin indices label residual components, and repeated indices are summed.

For a curve $\theta(t)$ parameterized by an affine parameter $t$, the geodesic equation with its first index lowered is
\begin{equation}
    g_{\mu\nu}\ddot{\theta}^{\nu}
    +
    \Gamma_{\mu,\alpha\beta}
    \dot{\theta}^{\alpha}\dot{\theta}^{\beta}
    =0,
    \label{eq:lower_index_geodesic}
\end{equation}
where dots denote derivatives with respect to the curve parameter $t$, and $\Gamma_{\mu,\alpha\beta}$ denotes the Christoffel symbol of the first kind. For the pullback metric, it is given by
\begin{align}
    \Gamma_{\mu,\alpha\beta}
    &=
    \frac{1}{2}
    \left(
        \partial_{\alpha}g_{\mu\beta}
        +
        \partial_{\beta}g_{\mu\alpha}
        -
        \partial_{\mu}g_{\alpha\beta}
    \right)
    \nonumber\\
    &=
    J_{m\mu}\,
    \partial_{\alpha}\partial_{\beta}r_m .
    \label{eq:christoffel}
\end{align}
Substituting into Eq.~\eqref{eq:lower_index_geodesic} gives
\begin{equation}
    J_{m\mu}
    \left(
        J_{m\nu}\ddot{\theta}^{\nu}
        +
        \partial_{\alpha}\partial_{\beta}r_m\,
        \dot{\theta}^{\alpha}\dot{\theta}^{\beta}
    \right)
    =0.
    \label{eq:embedded_geodesic_component}
\end{equation}

To interpret this expression, consider the residual trajectory $r\bigl(\theta(t)\bigr).$ By the chain rule, its second derivative with respect to $t$ is
\begin{equation}
    \ddot{r}_m(\theta(t))
    =
    J_{m\nu}\ddot{\theta}^{\nu}
    +
    \partial_{\alpha}\partial_{\beta}r_m\,
    \dot{\theta}^{\alpha}\dot{\theta}^{\beta}.
    \label{eq:residual_curve_acceleration}
\end{equation}
Equation~\eqref{eq:embedded_geodesic_component} is therefore equivalent to
\begin{equation}
    \mathcal{A}[\theta](t):=J\bigl(\theta(t)\bigr)^{\top}
    \ddot{r}\bigl(\theta(t)\bigr)
    =0.
    \label{eq:embedded_geodesic_matrix}
\end{equation}
We refer to $\mathcal A[\theta](t)$ as the geodesic residual of the parameter curve. Equation~\eqref{eq:embedded_geodesic_matrix} states that the residual acceleration is orthogonal to
$\operatorname{Col}\bigl(J(\theta(t))\bigr)$, the tangent space of the model manifold at $r\bigl(\theta(t)\bigr)$. Thus, a geodesic eliminates the tangential residual acceleration associated with the chosen parameterization while retaining the normal acceleration associated with the intrinsic curvature.

RNC-LM does not integrate Eq.~\eqref{eq:embedded_geodesic_matrix} directly. Instead, it constructs a finite-order curve whose geodesic residual vanishes successively in a Taylor expansion around \(t=0\), as described in the next subsection.

\subsection{Recursive construction of the RNC--LM coefficients}

We derive the coefficients \(c_2,\ldots,c_K\) of the curve
\(\theta_K(t)\) in Eq.~\eqref{eq:rnc_curve}, with \(c_1=v\) given by
the LM system in Eq.~\eqref{eq:sLM_eq}. Suppose that
\(c_1,\ldots,c_{n-1}\) are known and define the truncated curve
\begin{equation}
    \theta_{<n}(t)
    =
    \theta_p+
    \sum_{q=1}^{n-1}\frac{t^q}{q!}c_q.
    \label{eq:rnc_defect}
\end{equation}
Because the geodesic residual
\(\mathcal A[\theta](t)=J(\theta(t))^\top\ddot r(\theta(t))\)
is a second-order differential operator, \(c_n\) first contributes at
order \(t^{n-2}\). We therefore define the defect left by the truncated
curve as
\begin{equation}
    b_n
    :=
    \left.
    \frac{\mathrm d^{\,n-2}}{\mathrm dt^{\,n-2}}
    \mathcal A[\theta_{<n}](t)
    \right|_{t=0}
    .
    \label{eq:b_n_defect}
\end{equation}

Adding the unknown term
\(\delta\theta_n(t)=t^n c_n/n!\) gives
\(\theta_n(t)=\theta_{<n}(t)+\delta\theta_n(t)\). At order
\(t^{n-2}\), the only contribution from \(c_n\) enters through
\(\delta\ddot\theta_n=t^{n-2}c_n/(n-2)!\). Hence,
\begin{equation}
    \mathcal A[\theta_n](t)
    =
    \mathcal A[\theta_{<n}](t)
    +
    \frac{t^{n-2}}{(n-2)!}gc_n
    +
    O(t^{n-1}).
    \label{eq:rnc_perturbation}
\end{equation}
Exact cancellation would require $gc_n=-b_n$. RNC--LM applies the
same Marquardt regularization as in the initial LM step, yielding
\begin{equation}
    Gc_n=-b_n,
    \qquad n=2,\ldots,K.
    \label{eq:rnc_recursion2}
\end{equation}
All coefficient equations share the same matrix $G$. Its Cholesky factor provides both the local metric orthonormalization used by the RNC construction and the reusable factorization for every coefficient; a detailed derivation is given in the SI. 

For each recursion order $n$, define the derivatives along
$\theta_{<n}(t)$ gives
\begin{equation}
    R_k=
    \left.\frac{\mathrm d^k}{\mathrm dt^k}
    r(\theta_{<n}(t))\right|_{t=0},
    \qquad
    J_k=
    \left.\frac{\mathrm d^k}{\mathrm dt^k}
    J(\theta_{<n}(t))\right|_{t=0}.
\end{equation}
Applying the Leibniz rule to Eq.~\eqref{eq:b_n_defect} gives
\begin{equation}
    b_n
    =
    \sum_{k=0}^{n-2}
    \binom{n-2}{k}J_k^\top R_{n-k}
    =
    \sum_{k=0}^{n-2}
    \binom{n-2}{k}
    \left(\frac{\partial R_k}{\partial\theta_p}\right)^\top
    R_{n-k},
    \label{eq:bn_leibniz}
\end{equation}
where differentiation with respect to \(\theta_p\) holds
\(c_1,\ldots,c_{n-1}\) fixed. Taylor-mode automatic differentiation
generates \(R_0,\ldots,R_n\), and the sum is evaluated by a single
vector--Jacobian product (VJP) through \((R_0,\ldots,R_{n-2})\),
with \(\binom{n-2}{k}R_{n-k}\) treated as fixed cotangent weights.
Thus, each order requires a Taylor-mode evaluation, a VJP, and a
linear solve with the existing factorization of \(G\), without
explicitly constructing \(J_k\) or full higher-order derivative tensors.

The first two nontrivial orders clarify the geometric meaning
of the recursion:
\begin{align}
    Gc_2
    &=
    -J^\top R_2,
    \label{eq:second_order_coefficient}\\
    Gc_3
    &=
    -J^\top R_3-J_1^\top R_2,
    \label{eq:third_order_coefficient}
\end{align}
where the derivatives are evaluated along \(\theta_{<2}\) and
\(\theta_{<3}\), respectively.
Equation~\eqref{eq:second_order_coefficient} recovers geodesic
acceleration. In Eq.~\eqref{eq:third_order_coefficient}, the term
\(-J^\top R_3\) corresponds to the third-order residual-linear
correction~\cite{brooks2024higherordercorrectionsoptimisersbased},
whereas \(-J_1^\top R_2\) accounts for the variation of the tangent
space along the curve.

In the unregularized limit, the second-order equation ensures that
the residual acceleration \(R_2\) along \(\theta_{<3}\) is normal
at the expansion point, so \(J^\top R_2=0\). As the tangent space
moves, this fixed vector satisfies
\begin{equation}
    J(\theta_{<3}(t))^\top R_2
    =
    tJ_1^\top R_2+O(t^2).
\end{equation}
The additional third-order term compensates for this leading
tangential component, preserving the geodesic condition to the
next order. Explicit tensor forms and higher-order expressions
are provided in the SI.

In the absence of regularization, the recursion eliminates the Taylor
coefficients of the geodesic residual through order \(t^{K-2}\):
\begin{equation}
    \mathcal A[\theta_K](t)=O(t^{K-1}).
    \label{eq:rnc_approximation_order}
\end{equation}
At finite damping, the remaining defect is
\begin{equation}
    \mathcal A[\theta_K](t)
    =
    -\lambda D\ddot\theta_K(t)
    +
    O(t^{K-1}),
    \label{eq:damped_geometric_defect}
\end{equation}
which quantifies the deviation introduced by regularization.


\subsection{Tangential residual cancellation and LM-model consistency}

Now, let us briefly revisit the origin of the LM method; the LM subproblem is constructed from the linear approximation residual model.
\begin{equation}
    r(\theta_p + t c_1) \approx R_\mathrm{lin}(t)=r(\theta_p)+tJ(\theta_p) c_1
\end{equation}
The minimization of the corresponding regularized quadratic objective is precisely Eq.~\eqref{eq:sLM_eq}. By contrast, after the RNC curve is constructed, the real residual trajectory is $R_K(t) = r(\theta_K(t))$. Their difference is
\begin{equation}
    E_K(t)=R_K(t)- R_\mathrm{lin}(t),
\end{equation}

Under the local non-degeneracy assumption, define the orthogonal projector onto the moving tangent space by
\begin{equation}
    P_T(t)
    =
    J_K(t)
    \bigl[J_K(t)^\top J_K(t)\bigr]^{-1}
    J_K(t)^\top,
    \label{eq:moving_tangent_projector}
\end{equation}
where $J_K(t)=J(\theta_K(t)).$ For $\lambda\rightarrow0$, RNC curve satisfies Eq.~\eqref{eq:rnc_approximation_order}. Since \(\ddot{R}_\mathrm{lin}(t)=0\), it follows that
\begin{align}
    P_T(t)\ddot E_K(t)
    &=
    P_T(t)\ddot R_K(t)
    \nonumber\\
    &=
    J_K(t)
    \bigl[J_K(t)^\top J_K(t)\bigr]^{-1}
    \mathcal A[\theta_K](t)
    =
    O(t^{K-1}).
    \label{eq:tangential_error_cancellation}
\end{align}
Equivalently,
\begin{equation}
    \left.
    \frac{\mathrm d^\ell}{\mathrm dt^\ell}
    \left[P_T(t)\ddot E_K(t)\right]
    \right|_{t=0}
    =0,
    \qquad
    \ell=0,\ldots,K-2.
    \label{eq:tangential_cancellation_orders}
\end{equation}

Thus, RNC--LM eliminates the Taylor coefficients of the tangential acceleration of the linear residual-model error in the moving tangent frame. This is the precise sense in which the RNC trajectory suppresses the parameterization-dependent deviation from the LM linear residual model. The normal component, which reflects the intrinsic curvature of the model manifold, is not removed. At finite damping, this cancellation is perturbed by the regularization defect quantified in
Eq.~\eqref{eq:damped_geometric_defect}.

\subsection{Trust-ratio control of damping and curve length}


The damping parameter $\lambda$ and the curve parameter $t$ play distinct roles. For fixed $\lambda$, the LM direction and the coefficients $c_2,\ldots,c_K$ determine the shape of the RNC curve, while $t$ subsequently determines the distance travelled along this fixed curve.

Because RNC--LM leaves the LM subproblem unchanged, we retain the linear residual model
\begin{equation}
    m(t)
    =
    \frac{1}{2}\left\|r+tJc_1\right\|^2 .
    \label{eq:lm_curve_model}
\end{equation}
Following standard trust-ratio implementations of LM~\cite{more1978}, we compare the actual reduction along the RNC curve with the reduction predicted by the linear residual model. For a trial point $\theta_K(t)$, let $\phi(t):=C\bigl(\theta_K(t)\bigr)$, then actual and predicted reductions are $\Delta C(t)=\phi(0)-\phi(t)$, $\Delta m(t)=m(0)-m(t),$ and their ratio is
\begin{equation}
    \rho(t)
    =
    \frac{\Delta C(t)}{\Delta m(t)}.
    \label{eq:curve_trust_ratio}
\end{equation}
Using \(Gc_1 = -J^\top r\), the predicted reduction and the initial directional derivative of $\phi(0)$ are
\begin{equation}
    \Delta m(t) = (t - 0.5 t^2)\, c_1^\top G c_1 + 0.5\,\lambda\, t^2\, c_1^\top D c_1,
    \label{eq:curve_initial_slope}
\end{equation}
and
\begin{equation}
    \phi'(0) = -\sigma,
    \qquad
    \sigma := c_1^\top G c_1 .
\end{equation}

Each newly constructed curve is first evaluated at $t_1=1$. A trial is accepted when $\rho(t_j)>\eta_{\mathrm{acc}}$. If a trial $t_j$ is rejected, the observed value $\phi(t_j)$ still provides useful information about the objective along the curve that cannot be contained in $\theta_p$ alone. Together, with the known initial value $\phi(0)$, the exact directional derivative $\phi'(0)=-\sigma$, and the trial value $\phi(t_j)$ uniquely determine a quadratic interpolant~\cite{nocedal2006}:
\begin{equation}
    q_j(t)
    =
    \phi(0)-\sigma t+\left(\frac{\phi(t_j)-\phi(0)+\sigma t_j}{t_j^2}\right)t^2,
    \label{eq:curve_quadratic_model}
\end{equation}
Its minimizer is
\begin{equation}
    \widetilde t_{j+1}
    =
    \frac{\sigma t_j^2}
    {2[\phi(t_j)-\phi(0)+\sigma t_j]}.
    \label{eq:interpolated_curve_parameter}
\end{equation}
In practice, the interpolated value is restricted to a prescribed contraction interval, and provides the next trial curve parameter
\begin{equation}
    t_{j+1}
    =
    \operatorname{clip}
    \left(
        \widetilde t_{j+1},
        \beta_{\min}t_j,
        \beta_{\max}t_j
    \right),
    \qquad
    0<\beta_{\min}<\beta_{\max}<1.
    \label{eq:safeguarded_curve_parameter}
\end{equation}

For an accepted parameter \(t_{\mathrm{acc}}\), the iteration is updated as
\begin{equation}
    \theta_{p+1}=\theta_K(t_{\mathrm{acc}}).
\end{equation}
The damping parameter is then adjusted using the accepted trust ratio
$\rho_{\mathrm{acc}}=\rho(t_{\mathrm{acc}})$~\cite{more1978, PhysRevE_83_036701}
\begin{equation}
    \lambda_{\mathrm{next}}
    =
    \begin{cases}
        2\lambda,   & \rho_{\mathrm{acc}}<0.25,\\[2pt]
        \lambda/3, & \rho_{\mathrm{acc}}>0.75,\\[2pt]
        \lambda,   & \text{otherwise}.
    \end{cases}
    \label{eq:damping_update}
\end{equation}
If no acceptable curve parameter is found within the prescribed number of
trials, the update is rejected, \(\lambda\) is increased, and the RNC
coefficients are recomputed.

Changing \(t\) requires only a new residual and objective evaluation.
The Jacobian and the RNC coefficients remain unchanged throughout the inner curve search; the interpolation itself involves only
scalar operations.

\backmatter

\bibliography{sn-bibliography}

\end{document}